\documentclass[letterpaper]{article} 
\usepackage{aaai25}  
\usepackage{times}  
\usepackage{helvet}  
\usepackage{courier}  
\usepackage[hyphens]{url}  
\usepackage{graphicx} 
\usepackage{natbib}  
\usepackage{caption} 
\usepackage{algorithm}
\usepackage{algorithmicx}
\usepackage{amsmath}

\usepackage{algpseudocode}
\usepackage{subcaption}

\usepackage{capt-of,etoolbox}

\usepackage{newfloat}
\usepackage{booktabs}
\usepackage{multirow}
\usepackage{listings}
\DeclareCaptionStyle{ruled}{labelfont=normalfont,labelsep=colon,strut=off} 
\floatstyle{ruled}
\newfloat{listing}{tb}{lst}{}
\floatname{listing}{Listing}

\usepackage{appendix}

\title{Anywhere: A Multi-Agent Framework for User-Guided, Reliable, and Diverse Foreground-Conditioned Image Generation}
\author{
    Xie Tianyidan\textsuperscript{\rm 2},
    Rui Ma\textsuperscript{\rm 3},
    Qian Wang\textsuperscript{\rm 4}\thanks{Corresponding author},
    Xiaoqian Ye\textsuperscript{\rm 4},
    Feixuan Liu\textsuperscript{\rm 5},
    Ying Tai\textsuperscript{\rm 1,2},
    Zhenyu Zhang\textsuperscript{\rm 1,2},
    Lanjun Wang\textsuperscript{\rm 6},
    Zili Yi\textsuperscript{\rm 1,2}*%
}
\affiliations{%
    \textsuperscript{\rm 1}State Key Laboratory of Novel Software Technology, Nanjing University, Nanjing, China\\
    \textsuperscript{\rm 2}School of Intelligence Science and Technology, Nanjing University, Suzhou, China\\
    \textsuperscript{\rm 3}Jilin University, Changchun, China\\
    \textsuperscript{\rm 4}China Mobile Research Institute, Beijing, China\\
    \textsuperscript{\rm 5}Beijing Shuzhimei Technology Co., Ltd, Beijing, China\\
    \textsuperscript{\rm 6}School of New Media and Communication, Tianjin University, Tianjin, China\\
    sealical@outlook.com, yi@nju.edu.cn%
}

\usepackage{bibentry}

\begin{document}

\maketitle

\begin{abstract}
Recent advancements in image-conditioned image generation have demonstrated substantial progress. However, foreground-conditioned image generation remains underexplored, encountering challenges such as compromised object integrity, foreground-background inconsistencies, limited diversity, and reduced control flexibility. These challenges arise from current end-to-end inpainting models, which suffer from inaccurate training masks, limited foreground semantic understanding, data distribution biases, and inherent interference between visual and textual prompts. To overcome these limitations, we present Anywhere, a multi-agent framework that departs from the traditional end-to-end approach. In this framework, each agent is specialized in a distinct aspect, such as foreground understanding, diversity enhancement, object integrity protection, and textual prompt consistency. Our framework is further enhanced with the ability to incorporate optional user textual inputs, perform automated quality assessments, and initiate re-generation as needed. Comprehensive experiments demonstrate that this modular design effectively overcomes the limitations of existing end-to-end models, resulting in higher fidelity, quality, diversity and controllability in foreground-conditioned image generation. Additionally, the Anywhere framework is extensible, allowing it to benefit from future advancements in each individual agent.
\end{abstract}

\begin{figure}[t]
\centering
\includegraphics[width=\columnwidth]{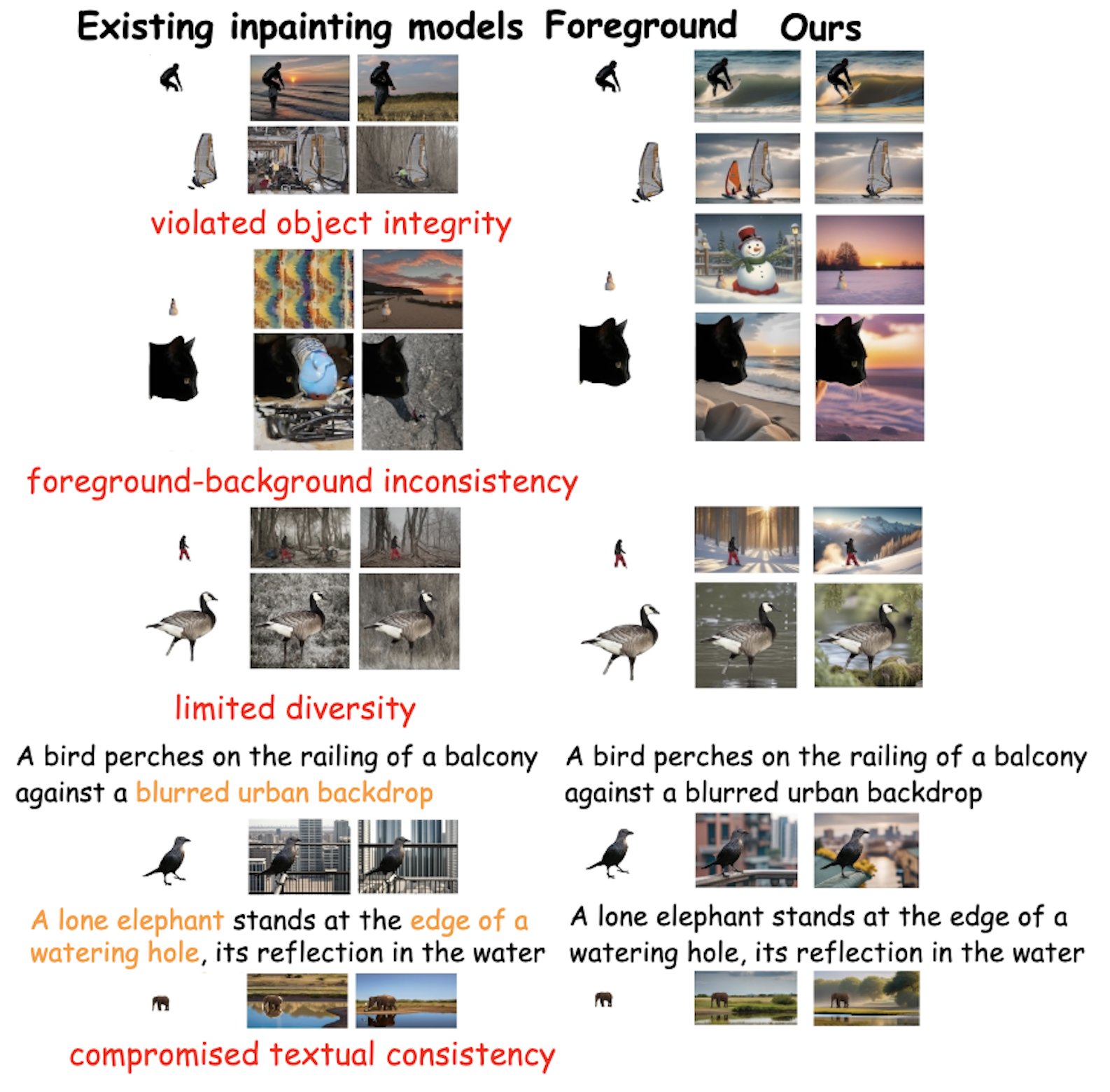}
\caption{Comparison of our Anywhere framework with inpainting models for foreground-conditioned image generation. The left section highlights the limitations of existing inpainting models, while the right section showcases our results. Our approach effectively addresses the issues (e.g., violated object integrity, foreground-background inconsistency, limited diversity, and compromised textual consistency), producing foreground-preserved, semantically coherent, diverse and text-consistent backgrounds tailored to the given foreground objects.}\label{fig:teaser}
\end{figure}

\section{Introduction}

Image generation conditioned on visual inputs has made remarkable strides in recent years, fueled by advancements in diffusion models \cite{ho2020denoising, zhang2023adding, huang2023composer, avrahami2023spatext, li2023gligen}. These models have enabled sophisticated techniques for tasks such as inpainting, image expansion, and object insertion \cite{rombach2022high, podell2023sdxl, manukyan2023hd, ju2024brushnet, zhang2024transparent}. However, the image generation task that attempts to complete the background based on a given foreground object remains underexplored. This technique enhances content creation, e-commerce visualization, and gaming by generating contextually appropriate backgrounds. Its significance lies in its wide-ranging applications, including virtual try-on, personalized advertising, and augmented reality. 

The complexity of foreground-conditioned image generation stems from its multifaceted nature, requiring simultaneous attention to object integrity, contextual relevance, and creative diversity. This task demands a deep understanding of visual semantics, spatial relationships, and creative composition. Existing inpainting methods often fail in three critical aspects: see Fig.~\ref{fig:teaser}. 

\begin{itemize}
    \item \textbf{Violated object integrity}: Existing inpainting methods often struggle to maintain the integrity of foreground objects, leading to the generation of unwanted elements or extensions. This issue arises because these methods rely on a single network to generate backgrounds in an end-to-end manner, heavily dependent on large datasets of foreground-background pairs obtained through auto-labeling with existing segmentation models or random masking. However, the accuracy of these segmentation masks is not always reliable, resulting in compromised object integrity and the inadvertent introduction of unwanted elements around the foreground object.
    \item \textbf{Foreground-background inconsistency}: Current image generation models, through trained to extend coherent structures from the input foreground, lack a deep semantic understanding of the foreground object and its relationship to the background, leading to contextually inappropriate or implausible scenes.
    \item \textbf{Limited diversity}: Existing models tend to generate monotonous or stereotypical backgrounds, failing to fully explore creative possibilities, as existing image generation models tend to incorporate and amplify biases from their training data (e.g., photographs with uniform or monotonous backgrounds).
    \item \textbf{Compromised Textual Consistency}: When applied to text-guided inpainting, the consistency of textual prompts can be compromised due to mutual interference that occurs during the joint integration of visual and textual inputs. This issue arises because text-guided inpainting models are typically adapted from text-to-image generators, which lack effective mechanisms and sufficient supervision to prevent unhealthy mutual interference between the visual and textual conditions.
\end{itemize}

Recognizing the limitations of end-to-end models, we propose a modular approach that incorporates multiple specialized agents to address the problem. Specifically, we introduce the Foreground Analyzer, based on advanced Visual Language Models (VLM) \cite{alayrac2022flamingo, li2022blip, achiam2023gpt, liu2024visual}, to achieve a deep understanding of foreground semantics. To enhance the diversity of generated images, the Prompt Creator leverages Large Language Models (LLM) \cite{touvron2023llama, achiam2023gpt} to produce creative textual prompts. In particular, we design the Template Repainter to protect object integrity by automatically detecting violations of object integrity and initiating repainting when necessary. Furthermore, the Quality Analyzer, also based on VLM, performs automated quality assessments and triggers regeneration if needed. In addition, our framework is extended to allow optional user textual inputs, composing prompts by merging these inputs with foreground semantics. 
 
We conducted comprehensive evaluations of our framework, demonstrating that the Anywhere framework significantly reduces instances of violated object integrity, improves foreground-background consistency, enhances diversity and controllability in foreground-conditioned image generation. Both subjective and objective metrics confirm that our approach excels in quality, diversity, user preference, and user controllability. In summary, the major contributions of this paper include:
\begin{itemize}
\item We introduce an innovative multi-agent framework specifically designed for foreground-conditioned image generation, representing a significant departure from traditional end-to-end models. This approach effectively overcomes the limitations of existing methods, leading to substantial improvements in the quality, robustness, diversity, and controllability of the generated results.

\item We design a Template Repainting Agent equipped with a unique mechanism for preserving object integrity and adaptive background synthesis. This agent successfully mitigates issues related to object integrity while maintaining contextual relevance, as validated by large-scale experiments.

\item Extensive evaluations reveal that our framework achieves a 4.6\% improvements in FID and an average 24\% increase in user preference scores, reduce 44\% of bad cases, along with a 33\% boost in the diversity score, compared to the best state-of-the-art inpainting models. In scenarios involving user textual inputs, our framework demonstrates a 5\% increase in text-image matching score over the leading text-guided image inpainting models.
\end{itemize}

\section{Related Works}
\subsection{Diffusion-based Controllable Image Generation}

Stable Diffusion, a leading text-to-image (T2I) model, has rapidly evolved beyond simple text inputs. While some researchers explore text-driven image-to-image generation \cite{hertz2022prompt, brooks2023instructpix2pix}, others have introduced diverse control signals to enhance the diffusion process. These include subject images \cite{gal2022image, ruiz2023dreambooth}, style information \cite{sohn2023styledrop}, layout conditions \cite{avrahami2023spatext}, edge maps \cite{zhang2023adding}, segmentation masks \cite{couairon2023zero}, and viewpoint control \cite{liu2023zero}. Notably, LayerDiffusion \cite{zhang2024transparent} generates images on transparent layers, allowing foreground or background elements to guide the process. These advancements demonstrate diffusion models' expanding capabilities to create more diverse and precise user-tailored images.

\subsection{Diffusion-based Image Inpainting}

Image inpainting is a pivotal task in computer vision, focusing on the restoration of masked regions based on surrounding unmasked content. Recent advancements in diffusion modeling have significantly propelled the field of inpainting forward. Notable techniques include Palette \cite{saharia2022palette} and Repaint \cite{lugmayr2022repaint}, which leverage the original image alongside the unmasked regions to enhance denoising. Blended Diffusion \cite{avrahami2022blended, avrahami2023blended} uses the known region to replace the unmasked region in the diffusion process. Additionally, Stable Diffusion Inpainting \cite{rombach2022high} introduces random masking during the text-to-image (T2I) process for training, augmented by supplementary textual inputs for precise control. Smartbrush \cite{xie2023smartbrush} exhibits the capability to tailor image results by manipulating mask types, while HD-Painter \cite{manukyan2023hd} and PowerPaint \cite{zhuang2023task} further refine the capabilities of SDI through additional training. BrushNet \cite{ju2024brushnet} stands out as a cutting-edge inpainting model, boasting plug-and-play functionality. Although these methods have yielded good results, there are still many difficulties in foreground-conditioned image generation, facing challenges such as violated object integrity where excessive content compromises foreground integrity, foreground-background inconsistency producing contextually inappropriate backgrounds, limited diversity and text-consistency in generated backgrounds. Hence more advanced approaches are needed for foreground-conditioned image generation.

\subsection{Large Language Model for Vision Task}

Natural language processing has undergone a dramatic transformation, with large language models (LLMs) approaching or surpassing human-level capabilities \cite{achiam2023gpt, touvron2023llama}. Simultaneously, visual question answering (VQA) has seen the emergence of high-performance models \cite{alayrac2022flamingo, li2022blip}. Despite high training costs impeding visual language model advancement, leveraging existing LLMs for visual tasks has become a key research direction \cite{brown2020language}. Models like LLaVA \cite{liu2024visual} and Bliva \cite{hu2024bliva} align LLMs with visual features, while others use LLMs as planners for visual tasks \cite{wu2023visual, gao2023assistgpt, shen2024hugginggpt, suris2023vipergpt}. Woodpecker \cite{yin2023woodpecker} and SIRI \cite{wang2023towards} enhance VLM reasoning through LLM knowledge. This trend reflects the growing application of large models to multi-modal tasks.

\begin{figure*}[h]
\centering
\includegraphics[width=\linewidth]{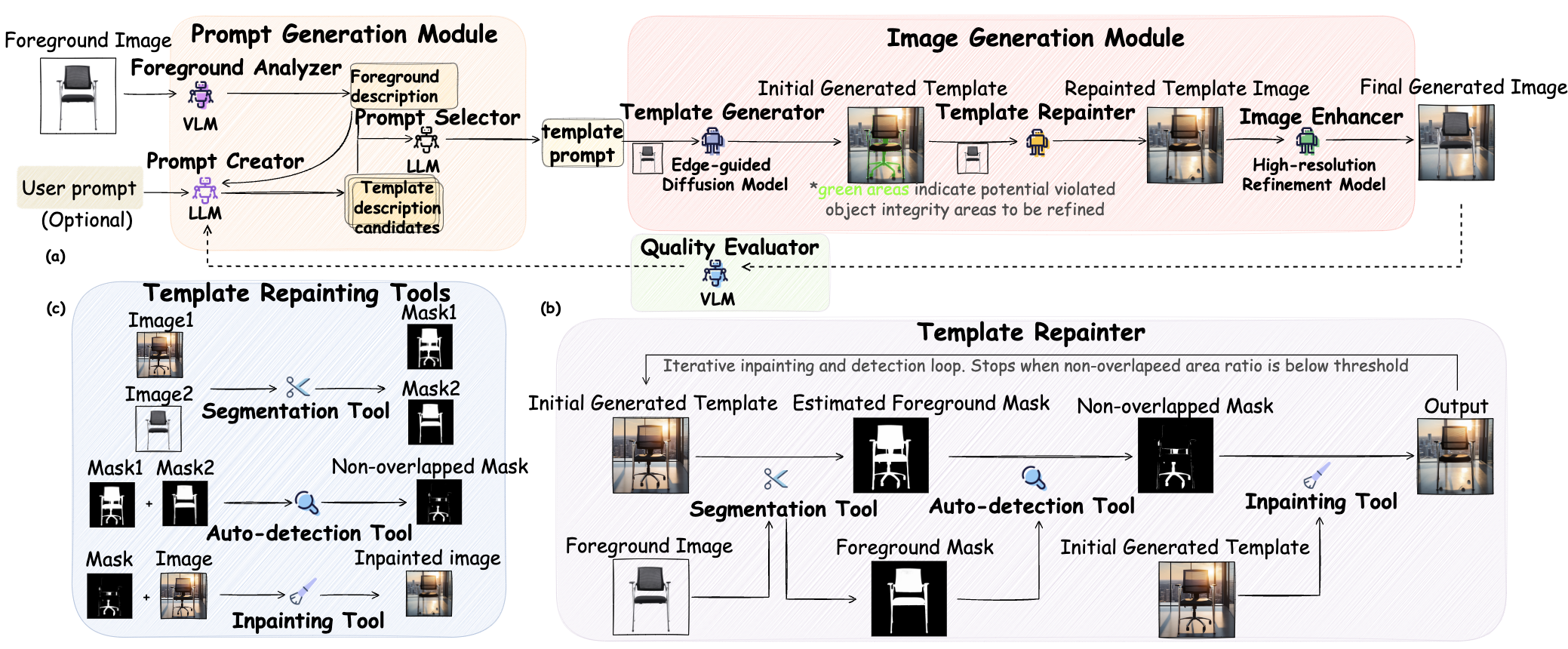}
\caption{Overview of the Anywhere framework. (a) Our approach comprises three main components: the Prompt Generation Module, the Image Generation Module, and the Quality Evaluator (Agent). The Prompt Generation Module uses a Foreground Analyzer (VLM) to extract textual descriptions from the foreground and a Prompt Creator (LLM) to generate multiple textual prompts based on the foreground descriptions and the user textual inputs if provided. The multiple textual prompts are then assessed by the Prompt Selector (LLM) and the best matched prompt will be selected. The Image Generation module includes a Template Generator (edge-guided image generation model) that generates a template image based on the textual prompt, a Template Repainter that detects object integrity violations (highlighted in green) and resolves the violations if needed, and an Image Enhancer (high-resolution image refinement Model) to paste-back the foreground and harmonize the final output. The Quality Evaluator Agent (VLM) assesses the resulting image, providing descriptive feedback and triggering re-generation when needed. (b) Illustration of the Template Repainter that performs violation detection by foreground segmentation and mask contrasting, and inpaints violated regions if they exist. (c) Illustration of template repainting tools used in the framework.}\label{fig:framework}
\end{figure*}

\section{Method}
\subsection{Framework Overview}
Anywhere is a multi-agent framework specifically designed to tackle the challenges of foreground-conditioned image generation. This framework integrates LLMs, VLMs, and image generation models into a sophisticated pipeline, as illustrated in Fig.~\ref{fig:framework} (a). The framework consists of three primary components: The Prompt Generation Module generates an elaborate textual prompt by leveraging the semantic understanding of the input foreground contents and the creative capabilities of LLMs; The Image Generation Module then utilizes the optimized textual prompt to create an appropriate template. The Quality Evaluator assesses the final image quality, providing descriptive and information-rich feedback to facilitate the re-generation of results.

\subsection{Prompt Generation Module}

The Prompt Generation Module employs specialized agents focused on foreground understanding and textual prompt optimization to address foreground-background inconsistencies, incorporate optional user inputs, enhance text prompt consistency, and boost diversity. The key agents in this module are:

\subsubsection{Foreground Analyzer}
This VLM-based agent extracts detailed textual information from the foreground image, capturing rich attributes of the foreground object, such as object type, shape, color, pose, and material. A comprehensive list of template questions guides the extraction of these specific attributes. The agent outputs finely detailed textual information in a structured, JSON-formatted description.

\subsubsection{Prompt Creator}
This LLM-based agent generates diverse textual descriptions based on extracted foreground details and, when available, integrates user input and descriptive feedback. Acting as a creative engine, the agent explores imaginative compositions to enhance output diversity. It utilizes a pre-designed prompting scheme that ensures generated texts are both varied and compatible with image generation models. The agent combines three distinct inputs: foreground textual descriptions, user input if provided, and textual feedback from the Quality Evaluator, as illustrated in Fig. \ref{fig:framework}. By merging information from these sources, the agent generates k candidate textual prompts.

\subsubsection{Prompt Selector}
This LLM-based agent evaluates the prompts generated by the Prompt Creator, taking into account factors such as relevance to the foreground details and compatibility with image generation models. It ranks the prompts based on evaluation scores, with the top-ranked prompts being selected probabilistically.

The Prompt Generation Module is designed to create textual prompts that facilitate the subsequent template generation process, ensuring relevance to the foreground, visual diversity, and text consistency. Additional details about this module are provided in the Appendix.

\subsection{Image Generation Module}

The Image Generation Module represents a major advancement in foreground-conditioned image generation by transforming template prompts into visually compelling backgrounds that seamlessly integrate with foreground images. Our multi-step scheme effectively addresses key challenges, particularly object integrity and foreground-background consistency. This is achieved through the use of three specialized agents that provide unparalleled control and precision throughout the generation process:

\subsubsection{Template Generator}
We utilize ControlNet \cite{zhang2023adding}, an edge-guided image generation model, to create initial background templates. This innovative application of ControlNet ensures spatial coherence between the generated backgrounds and the foreground objects, maintaining fine-grained control while producing high-quality images. The generation process is grounded in both the edge map of the foreground image and the template prompt, establishing a solid foundation for contextually appropriate backgrounds.

\subsubsection{Template Repainter} This targeted agent allows for precise and efficient corrections, preserving object integrity and ensuring consistent foreground-background integration, as illustrated in Fig.~\ref{fig:framework} (b). It operates through the following key components (depicted in Fig.~\ref{fig:framework} (c)):

\begin{itemize} \item \textbf{Segmentation Tool.} This tool generates an estimated foreground mask from the initial template image, which is further used by the auto-detection tool. 
\item \textbf{Auto-detection Tool.} This tool uses the estimated foreground mask and the actual foreground mask to detect areas where object integrity is compromised. It first identifies the foreground bounding box with detection model \cite{liu2023grounding} in the input image, then utilizes the bounding box to crop both masked images, and finally calculates a non-overlapped mask by comparing the cropped estimated and actual masks, pinpointing regions requiring repainting. Note that repainting is not triggered if the non-overlapped area is below a certain threshold.
\item \textbf{Inpainting Tool}. This advanced model selectively inpaints the areas identified by the non-overlapped mask, generating contextually appropriate content.
\end{itemize}

\subsubsection{Image Enhancer} Powered by a high-resolution refinement model (such as Stable-Diffusion XL \cite{podell2023sdxl}), this agent enhances the overall quality of the composite image, focusing on fine details, color balance, and smooth transitions.

The collaborative effort of these agents transforms the template prompt and foreground image into a contextually appropriate and visually compelling final image. This process effectively realizes the creative vision established in the Prompt Generation Module while addressing the unique challenges of foreground-conditioned image generation. More detailed information about the Image Generation algorithm can be found in the Appendix.

\subsection{Quality Evaluator}
The Quality Evaluator, a VLM-based agent, enhances the final image quality through a feedback loop. This agent leverages the VLM's advanced capabilities to evaluate visual relationships, content rationality, and overall image quality, providing a nuanced and comprehensive assessment that surpasses traditional image quality metrics \cite{achiam2023gpt, team2023gemini, li2022blip}. We have developed a detailed questionnaire prompt list to facilitate the analysis of foreground-background integration, focusing on factors such as lighting consistency, color harmony, structural coherence, spatial relationships, and semantic relevance.

Based on this analysis, the Quality Evaluator generates detailed textual feedback, which is communicated to the Prompt Generation Module for potential re-generation. This feedback loop significantly enhances the system's ability to create reliable and high-fidelity compositions. To prevent endless generation iterations, the system is capped at a maximum of three loops, as validated by experimental results (see Appendix for more details).

\section{Experiments}
\subsection{Experimental Setup} \label{sec:exp_setup}
\subsubsection{Dataset} 
We curated a foreground dataset consisting of 3,000 images by randomly selecting 1,500 images from the LAION dataset \cite{schuhmann2022laion} and 1,500 from the MSCOCO dataset \cite{lin2014microsoft}. Each original image was segmented, and a randomly chosen foreground segment was extracted to serve as the test data. This approach ensures a diverse range of in-the-wild scenarios and object types (e.g., humans, vehicles, pets).

\subsubsection{Implementation Details}

Our framework integrates state-of-the-art models across its various components. The Prompt Generation Module leverages Gemini-Pro (LLM) \cite{team2023gemini} for both the Prompt Creator and Prompt Selector, and Gemini-Pro-Vision (VLM) \cite{team2023gemini} for the Foreground Analyzer. In the Image Generation Module, we employ ControlNet\_SDXL\_Canny \cite{cnxlcanny} as the Template Generator and SDXL Refiner \cite{sdxlrefiner} as the Image Enhancer. The Template Refinement Tools consist of RMBG-1.4 \cite{rmbg} for segmentation, Grounding DINO \cite{liu2023grounding} for auto-detection, and LaMa \cite{suvorov2022resolution} for inpainting. Additionally, the Quality Evaluator uses Gemini-Pro-Vision (VLM) \cite{team2023gemini}. Detailed prompt templates for the LLM and VLM components are provided in the Appendix.

\subsubsection{Baseline}\label{sec:baseline}

To assess the effectiveness of our proposed framework, we compared it with three state-of-the-art inpainting models: BrushNet \cite{ju2024brushnet}, a plug-and-play dual-branch model for image inpainting; HD-Painter \cite{manukyan2023hd}, a high-resolution inpainting model known for precise prompt adherence; and Stable Diffusion 2.0 Inpainting \cite{rombach2022high, sd2inpainting}, a widely-used diffusion-based inpainting model. These methods represent the cutting edge in image inpainting.

\subsubsection{Evaluation Metrics}\label{sec:metrics}

We employ a comprehensive set of metrics to evaluate the aesthetic quality, human preference alignment, and image fidelity of the generated images. The Aesthetic Score (AS) \cite{schuhmann2022laion} assesses overall visual appeal, while PickScore \cite{kirstain2023pick} simulates human rating behavior. The Human Preference Score (HPS) \cite{wu2023human} directly measures human preference in comparison to real images, and the CLIP Similarity score (CLIP-Sim) \cite{radford2021learning} quantifies the semantic alignment between generated images and input text prompts, employing ViT-B/16 as its image encoder. Additionally, we use Fréchet Inception Distance (FID) \cite{heusel2017gans} to evaluate image naturalism and fidelity. This selection of metrics offers a thorough assessment of our framework's performance across aesthetic quality, human preference alignment, text-image coherence, image diversity, and fidelity.

\subsection{Qualitative Results}

\begin{figure*}[th]
\centering
\includegraphics[width=\linewidth]{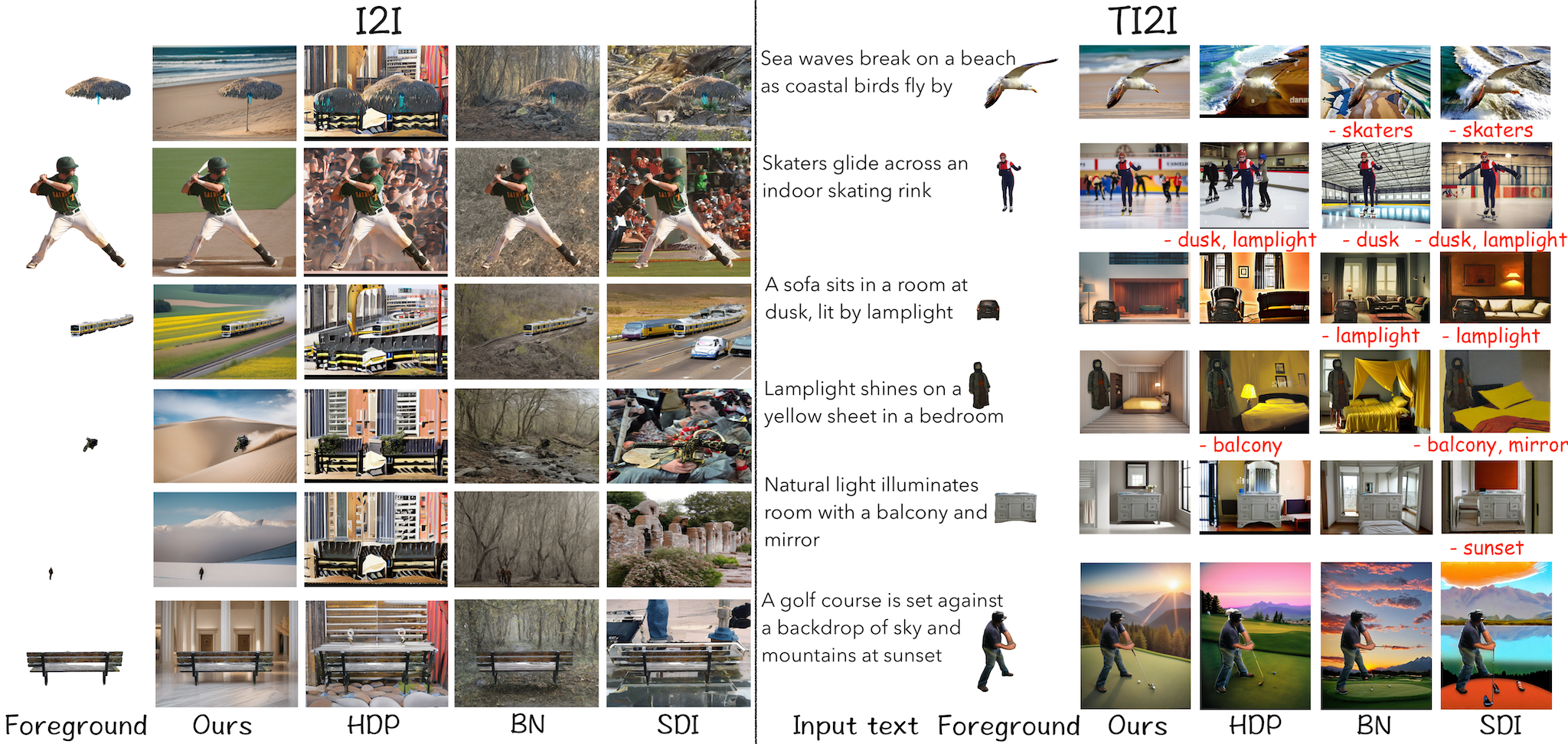}
\caption{We compare our approach to advanced inpainting models on foreground-conditioned image generation tasks in both text-free (I2I) and text-guided (TI2I) scenarios. These results are generated using unconstrained, in-the-wild foreground images. Red color indicates missing elements in generated images. The inpainting models used for comparison include HD-Painter (\textbf{HDP}), BrushNet (\textbf{BN}), and Stable Diffusion 2.0 Inpainting (\textbf{SDI}). }\label{fig:qual_pipe}
\end{figure*}

The qualitative comparison results are presented in Fig.~\ref{fig:qual_pipe}. In the text-free scenario, our approach generates contextually appropriate and diverse backgrounds while maintaining object integrity. In contrast, HD-Painter (HDP) and Stable Diffusion Inpainting (SDI) often produce inconsistent or illogical backgrounds and compromise object integrity. BrushNet (BN) tends to generate uniform backgrounds with limited creativity. In the text-guided scenario, our framework effectively integrates user text input to create coherent backgrounds that seamlessly blend with the foreground and prompt. HDP frequently violates object integrity (see rows 1-3 and 6) and overlooks key textual elements (see row 3). BN attempts to incorporate prompts but often places foregrounds in unsuitable settings (see rows 1-3 and 6), misses textual elements (see rows 2-4), and compromises object integrity (see row 4). SDI struggles to align its outputs with the given prompts, leading to irrelevant backgrounds (see rows 2-6).

\begin{table*}[t]
    \centering
    \resizebox{0.65\textwidth}{!}{%
    \begin{tabular}{llccccc}
    \toprule
        & & \multicolumn{3}{c}{Aesthetic and Human Preference} & \multicolumn{1}{c}{Text Align} & \multicolumn{1}{c}{Image Fidelity} \\
    \cmidrule(lr){3-5}\cmidrule(lr){6-6}\cmidrule(lr){7-7}
        Task & Method & AS ($\uparrow$) & PickScore ($\uparrow$) & HPS ($\uparrow$) & CLIP-Sim ($\uparrow$) & FID ($\downarrow$) \\
    \midrule
        \multirow{4}{*}{\rotatebox[origin=c]{90}{I2I}} 
        & BrushNet      & 5.238 & 0.261 & 0.186 & - & 41.047 \\
        & HD-Painter    & 4.899 & 0.133 & 0.178 & - & 38.797 \\
        & SD Inpainting & 4.988 & 0.186 & 0.180 & - & 42.342 \\
        & Ours  & \textbf{5.604} & \textbf{0.418} & \textbf{0.190} & - & \textbf{37.007} \\
    \midrule
        \multirow{4}{*}{\rotatebox[origin=c]{90}{TI2I}}
        & BrushNet      & 5.467 & 0.302 & 0.184 & 17.483 & 38.313  \\
        & HD-Painter    & 4.971 & 0.126 & 0.173 & 17.466 & 37.398 \\
        & SD Inpainting & 5.132 & 0.177 & 0.175 & 17.314 & 39.476 \\
        & Ours & \textbf{5.812} & \textbf{0.394} & \textbf{0.195} & \textbf{18.357} & \textbf{36.450} \\
    \bottomrule
    \end{tabular}
    }
    \caption{Quantitative comparisons of our framework with advanced inpainting models on foreground-conditioned image generation tasks in both text-free (I2I) and text-guided (TI2I) scenarios.}
    \label{tab:quantitative_results}
\end{table*}

\subsection{Quantitative Results}
We conducted quantitative experiments for both text-free (Image-to-Image, or I2I) and text-guided (Text-guided Image-to-Image, or TI2I) scenarios using the metrics described in Sec.~\ref{sec:exp_setup}. The comparative results are presented in Tab.~\ref{tab:quantitative_results}.

\subsubsection{Metric Calculation Details}

Our evaluation methodology generates one resulting image per model for each test case. The assessment process is structured as follows: Aesthetic Score (AS) is directly calculated from the images produced by each model, without the need for additional textual information. For metrics that require textual input (PickScore, HPS, IR), the text-free (I2I) scenarios use the foreground object type name extracted by the Foreground Analyzer as the textual input for evaluation. For instance, in Fig.~\ref{fig:framework}, the foreground type is ``chair''. These metrics are then calculated for each generated image using the corresponding foreground label. In the text-guided (TI2I) scenarios, we conducted experiments with two types of user inputs: (1) generic phrases that cover both indoor and outdoor natural scenes, such as ``sunset'', ``snow'', ``room'', and ``beach''; and (2) unique sentences tailored to each foreground image, generated using a VLM. The results presented for TI2I are the average scores across both the generic phrases and the VLM-generated unique sentences.

\subsubsection{Results Analysis}
As shown, our framework consistently outperforms the state-of-the-art inpainting models across all metrics in both text-free (I2I) and text-guided (TI2I) scenarios. Our method achieves the highest scores in aesthetic quality and human preference, indicating superior visual appeal. In the text-guided scenario, our approach also shows significant improvements in text alignment, highlighting its effectiveness in enhancing textual consistency. While slight trade-offs are observed in some metrics, these reflect the inherent challenges of balancing foreground compatibility with prompt adherence. Notably, our framework consistently achieves the lowest FID scores across both tasks. These quantitative results align with our qualitative findings, confirming that the Anywhere multi-agent framework excels at generating visually appealing, diverse backgrounds while maintaining high relevance to both foreground objects and text prompts.

\subsection{User Study}

To validate our quantitative findings and assess real-world user preferences, we conducted a comprehensive user study involving 10 participants. The study evaluated a total of 100 foreground images, evenly divided between text-free (I2I) and text-guided (TI2I) scenarios for the foreground-conditioned image generation task. For the text-guided scenario, we randomly assigned prompts from a fixed list of textual templates to each image. Each test case was processed by our Anywhere framework and three state-of-the-art models: BrushNet \cite{ju2024brushnet}, HD-Painter \cite{manukyan2023hd}, and Stable Diffusion 2.0 Inpainting \cite{podell2023sdxl}. To ensure a thorough evaluation, each comparative method generated 3 results per test case, resulting in a total of 1,200 images (100 test cases $\times$ 4 methods $\times$ 3 results per method). Participants were asked to assess three key aspects of the generated images: aesthetic quality (rated on a scale of 1-5, with 5 indicating the highest quality), diversity (rated on a scale of 1-3, with 3 indicating the highest level of diversity), and the identification of any bad cases they deemed unsatisfactory or problematic (e.g., object integrity violations, illogical content, severe artifacts, etc.). After collecting the ratings, The ratings for each method were averaged separately across the three evaluated aspects.

\begin{table}[th]
\centering
\resizebox{0.9\columnwidth}{!}{%
\begin{tabular}{lccc}
\toprule
Method  & \begin{tabular}[c]{@{}c@{}}Aesthetic\\Quality ($\uparrow$)\end{tabular} & \begin{tabular}[c]{@{}c@{}}Diversity\\Score ($\uparrow$)\end{tabular} & \begin{tabular}[c]{@{}c@{}}Bad Case\\Rate ($\downarrow$)\end{tabular} \\
\midrule
BrushNet      & 2.90 & 1.93 & 0.34 \\
HD-Painter    & 1.86 & 1.78 & 0.66 \\
SD Inpainting & 2.12 & 1.85 & 0.47 \\
\midrule
Ours  & \textbf{3.45} & \textbf{2.57} & \textbf{0.19} \\
\bottomrule
\end{tabular}
}
\caption{The user studies that compare our approach with advanced inpainting models by evaluating the aesthetics, diversity, and rate of bad cases in the generated results.}
\label{tab:user_study}
\end{table}

\begin{figure}[t]
\centering
\includegraphics[width=0.85\columnwidth]{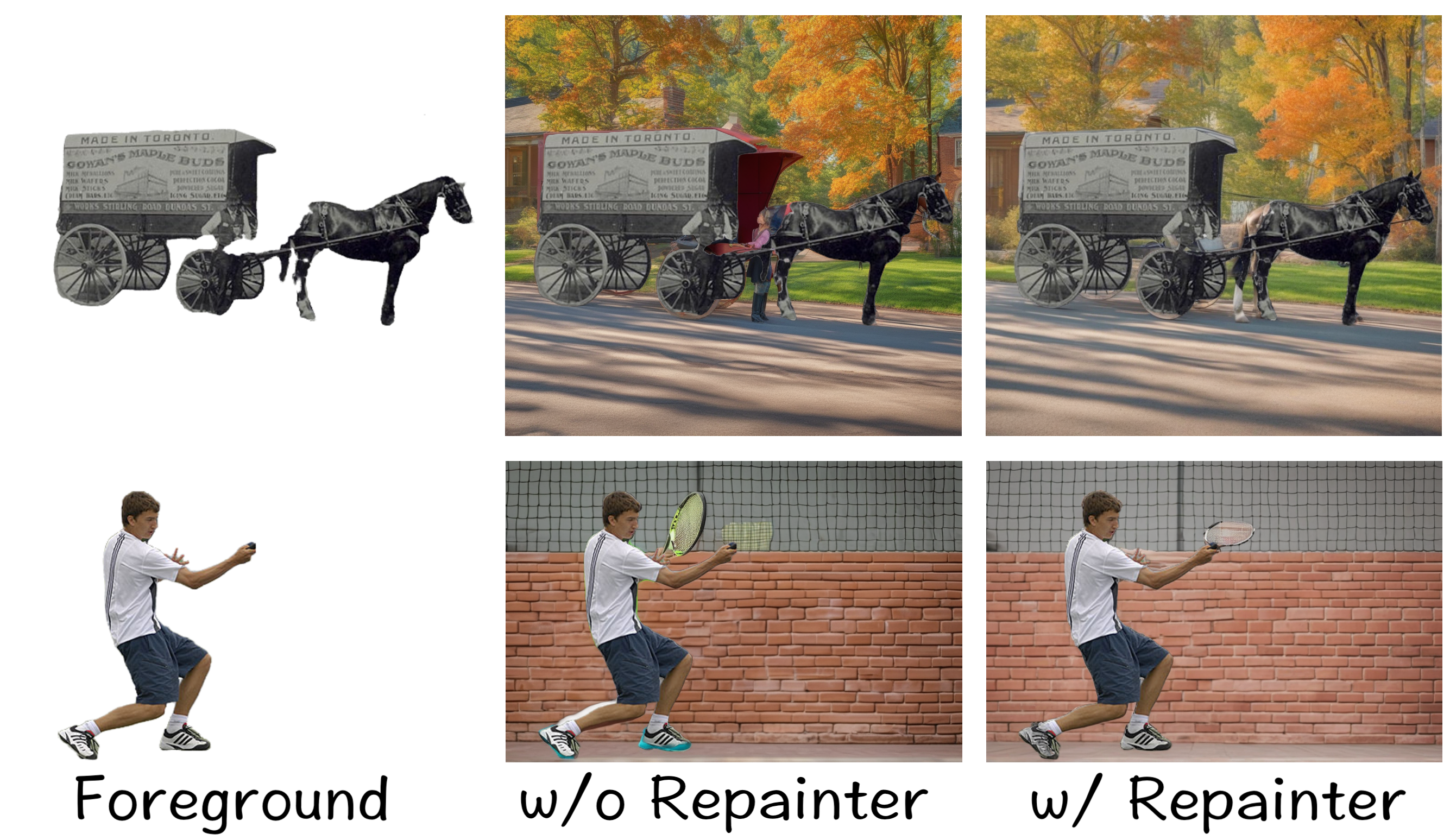}
\caption{Ablation studies on the Template Repainter.}\label{fig:ablation_repainter}
\end{figure}

As shown in Tab.~\ref{tab:user_study}, our Anywhere framework consistently outperformed existing methods across all evaluated metrics. It achieved the highest aesthetic quality score, marking a significant improvement over the next best performer. The diversity score of our method substantially surpassed that of BrushNet, highlighting our framework's ability to generate a broader range of creative and contextually appropriate backgrounds. Additionally, our method exhibited the lowest bad case rate, demonstrating considerable improvements over both BrushNet and Stable Diffusion Inpainting. These results strongly validate the effectiveness of our framework in producing high-quality, diverse, and reliable outputs.

\begin{table}[t]
    \centering
    \resizebox{\columnwidth}{!}{%
    \begin{tabular}{llccc}
    \toprule
        Task & Component Variant & AS ($\uparrow$) & CLIP-Sim ($\uparrow$) & FID ($\downarrow$) \\
    \midrule
        \multirow{6}{*}{\rotatebox[origin=c]{90}{I2I}}
        &    w/o PGM    &    5.263    &    -   & 40.751 \\
        &    w/o TR     &    5.417    &    -   & 38.833 \\
        &    w/o IE     &    5.518    &    -   & 37.980 \\
        &    w/o FA     &    5.395    &    -   & 38.968 \\
        &    w/o QE     &    5.462    &    -   & 38.510 \\
        & \textbf{Ours} & \textbf{5.604} & -   & \textbf{37.007} \\
    \midrule
        \multirow{4}{*}{\rotatebox[origin=c]{90}{TI2I}}
        &    w/o PC     &     5.561     &      18.077      & 37.346 \\
        &    w/o PS     &     5.624     &      18.165      & 37.031 \\
        &    w/o QE     &     5.537     &      18.211      & 36.952 \\
        & \textbf{Ours} & \textbf{5.812} & \textbf{18.357} & \textbf{36.450} \\
    \bottomrule
    \end{tabular}%
    }
\caption{Ablation studies for the Anywhere framework. PGM: Prompt Generation Module, TR: Template Repainter, IE: Image Enhancer, FA: Foreground Analyzer, QE: Quality Evaluator, PC: Prompt Creator, PS: Prompt Selector.}
\label{tab:ablation_results}
\end{table}

\subsection{Ablation Study}

For the ablation study, we systematically removed or modified various modules and assessed their impact on the evaluation metrics. In the ablation studies of the Prompt Generation Module, we directly set the user input text (if available) as the textual prompt for the Image Generation Module; if user input was not provided, an empty textual prompt was used instead. Without the Template Repainter, the initial template image was used as-is, without any additional processing. In the ablation studies of the Image Enhancer, we simply pasted the foreground object back onto the Template Repainter's output to produce the final result. To ablate the Foreground Analyzer, the foreground descriptions were excluded from the Prompt Creator's process. Without the Prompt Creator, we used the foreground descriptions concatenated with the user input text (if provided) as candidate template prompts for selection. For the Prompt Selector ablation, we randomly selected one of the outputs from the Prompt Creator as the template prompt.

Results in Tab.~\ref{tab:ablation_results} demonstrate the impact of these ablation studies. The removal of the Prompt Generation Module resulted in significant performance drops in FID and aesthetics scores, underscoring its crucial role in enhancing quality and diversity. The absence of the Foreground Analyzer led to a greater decline in quality than the removal of the Prompt Creator or Prompt Selector, although the latter two are vital for maintaining textual consistency in text-guided scenarios. Both the Template Repainter and the Image Enhancer had a notable impact on image quality metrics. The Quality Evaluator, while contributing less to overall quality and text consistency, still played a role. The qualitative results in Fig.~\ref{fig:ablation_repainter} highlight the importance of the Template Repainter in resolving issues of object integrity. Additional qualitative ablation results are presented in the Appendix.

\section{Conclusion and Future Work}

In this paper, we present Anywhere, a novel multi-agent framework for foreground-conditioned image generation that significantly outperforms existing end-to-end models in reliability, quality, diversity, and controllability. Our modular design, incorporating advanced VLMs, LLMs, and image generation models, effectively addresses critical challenges such as object integrity violations, foreground-background inconsistencies, limited diversity, and textual prompt inconsistencies. Our framework demonstrates substantial improvements over leading end-to-end inpainting models, with gains of 4.6\% in FID, 24\% in average human preference score, 33\% in diversity score, and 5\% in text-image matching score, while reducing bad cases by 44\%.

However, these advancements come at the cost of increased computational demands, requiring approximately 2$\sim$3$\times$ more GPU time than current end-to-end models on average for each generation. Additionally, the framework struggles with corner cases, such as transparent foreground objects. Future work will focus on optimizing computational efficiency and developing novel techniques to better handle long-tail scenarios.

\section*{Acknowledgments}
This work was supported by the National Natural Science Foundation of China (Grant No. 62406134, No. 62202199) and the Nanjing University-China Mobile Communications Group Co. Ltd. Joint Institute.

\bibliography{aaai25}
\appendix
\clearpage  
\onecolumn  

\begin{appendices}

\section{Appendix A: Prompt Templates of Anywhere Framework Components}

Our framework incorporates various prompt templates for different components, as illustrated in Fig. 2. These templates are crucial for guiding the behavior of our Large Language Models (LLMs) and Vision Language Models (VLMs) throughout the Image Generation process.

\subsection{Foreground Analyzer (VLM)}
The Foreground Analyzer extracts detailed textual information from the foreground image. The prompt template for this VLM-based agent is:

\begin{lstlisting}[numbers=none, frame=single, breaklines=true]
You are an expert analyst and observer. Please provide a detailed description of the given image, highlighting its important features. Include the name of the main object, color, materials, and its viewpoint. Structure your response in JSON format as follows: {'description':'', 'viewpoint':'', 'color':'', 'object_name':''}.
\end{lstlisting}

\subsection{Prompt Creator (LLM)}
The Prompt Creator generates diverse textual descriptions based on extracted foreground details and, when available, integrates user input and descriptive feedback. The prompt template for this LLM is:

\begin{lstlisting}[numbers=none, frame=single, breaklines=true]
As an imaginative photographer, I'll provide some essential information about this object (object name, viewpoint, color, and its description): [{object_name}], [{viewpoint}], [{color}] and [{description}]. The feedback about the object's previous scene description is: [{feedback}].
The user provided scene keywords [{user_prompt}].
Please generate 5 sets of relevant scene descriptions of this object incorporating the user scene keywords if they're not null; otherwise, provide 5 sets of relevant scene descriptions of this object based on its characteristics. Provide your rankings in JSON format: {'scene_descs':['scene1','scene2',...]}
\end{lstlisting}

\subsection{Prompt Selector (LLM)}
The Prompt Selector evaluates the prompts generated by the Prompt Creator, taking into account factors such as relevance to the foreground details and compatibility with image generation models. The prompt template is:

\begin{lstlisting}[numbers=none, frame=single, breaklines=true]
As an expert analyst, assess the correlation between the object description [{description}] and these 5 scene descriptions: [{scene_descs}]. Rank them from 1 to 5 based on appropriateness. Provide your rankings in JSON format: {'scene1':'rank_score1', 'scene2':'rank_score2'}.
\end{lstlisting}

\subsection{Quality Evaluator (VLM)}
The Quality Evaluator enhances the final image quality through a feedback loop. The prompt template is:

\begin{lstlisting}[numbers=none, frame=single, breaklines=true]
As a meticulous visual analyst, please address the following queries regarding the generated image: Is it typical for the [{object_name}] to be situated in this context? Does the [{object_name}] appear to be positioned realistically on a surface or the ground? Present your comprehensive analysis in JSON format: {'feedback':''}.
\end{lstlisting}

\subsection{Unique sentences generator (VLM)}
This unique sentences generator, elaborated upon in the Metric Calculation Details of Section 4.3, produces unique sentences specifically crafted for each foreground image (used in text-guide scenarios(TI2I)). The prompt template is as follows:

\begin{lstlisting}[numbers=none, frame=single, breaklines=true]
Envision the essential elements typically present in this scene (excluding the foreground), and synthesize these key components into a cohesive sentence. Provide your answer in JSON format: {'foreground_sentence':''}.
\end{lstlisting}

\section{Appendix B: Algorithms of Anywhere Framework Key Components}

\subsection{Prompt Generation module}
The algorithm for Prompt Generation module of the Anywhere framework, as illustrated in Figure 2 (a), is presented below: 

\begin{algorithm}
\caption{Prompt Generation Process}\label{alg:prompt_gen}
\begin{algorithmic}[1]
\Require Foreground image $I$, user prompt $
P_u$ (optional), feedback $P_{fb}$ (if provided), Foreground Analyzer (VLM) $F_{FA}(\cdot)$, Prompt Creator (LLM) $F_{PC}(\cdot,\cdot)$, Prompt Selector (LLM) $F_{PS}(\cdot,\cdot)$
\If{$P_u$ is not null}
    \State $P_u \gets P_u$ \Comment{Incorporate user prompt}
\Else
    \State $P_u \gets \emptyset$ \Comment{Empty set if no user prompt}
\EndIf
\If{$P_bf$ is not null}
    \State $P_{bf} \gets P_{bf}$ \Comment{Incorporate feedback prompt}
\Else
    \State $P_{bf} \gets \emptyset$ \Comment{Empty set if no feedback prompt}
\EndIf
\State $D \gets F_{FA}(I)$ \Comment{Generate structured foreground description}
\State $C \gets F_{PC}(D, P_u, P_{bf})$ \Comment{Generate set of template description candidates}
\State $T \gets F_{PS}(D, C)$ \Comment{Rank candidates and select top template prompt}
\State \Return $T$ \Comment{Return the optimal template prompt}
\end{algorithmic}
\end{algorithm}

\subsection{Image Generation module}
The algorithm for Image Generation module of the Anywhere framework, as illustrated in Figure 2 (a), is presented below: 

\begin{algorithm}
\caption{Image Generation Process}
\label{alg:image_generation}
\begin{algorithmic}[1]
\Require Foreground image $I$, template prompt $P$, Template Generator $G_{TG}(\cdot,\cdot)$, Template Repainter $G_{TR}$, Image Enhancer $G_{IE}(\cdot)$
\State $I_e \gets \text{EdgeExtraction}(I)$ \Comment{Extract edge map from foreground}
\State $T \gets G_{TG}(P, I_e)$ \Comment{Generate initial template image}
\State $I_r \gets G_{TR}(I, T)$ \Comment{Repaitning violated object integrity areas}
\State $I_c \gets \text{CompositeImage}(I_r, I)$ \Comment{Composite foreground onto repainted template image}
\State $I_{final} \gets G_{IE}(I_c)$ \Comment{Enhance the composite image}
\State \Return $I_{final}$ \Comment{Return the final generated image}
\end{algorithmic}
\end{algorithm}

\subsection{Template Repainter}
The algorithm for Template Repainter agent of the Anywhere framework, as depicted in Figure 2 (b) and (c), is presented below. The threshold $\theta$ is set to 0.03, representing the maximum acceptable ratio of non-overlapped area to the foreground mask area. The $max\_iter$ is set to 2, limiting the number of refinement iterations to balance efficiency and quality.

\begin{algorithm}
\caption{Template Repainter Process}
\label{alg:template_repainter}
\begin{algorithmic}[1]
    \Require Initial Generated Template $T$, Foreground Image $I$, Segmentation Tool $S(\cdot)$, Auto-detection Tool $A(\cdot,\cdot)$, Inpainting Tool $P(\cdot,\cdot)$, Threshold $\theta$, Max Iterations $max\_iter$
    \State $T_r \gets T$ \Comment{Initialize repainted template}
    \State $iter \gets 0$ \Comment{Initialize iteration counter}
    \Repeat
    \State $M_{ef} \gets S(T_r)$ \Comment{Create estimated foreground mask from template}
    \State $M_f \gets S(I)$ \Comment{Create foreground mask from input image}
    \State $M_{no} \gets A(M_{ef}, M_f)$ \Comment{Generate non-overlapped mask}
    \State $ratio \gets \text{CalculateNonOverlapRatio}(M_{no}, M_f)$
    \If{$ratio > \theta$}
        \State $T_r \gets P(T_r, M_{no})$ \Comment{Apply inpainting to violated areas}
    \EndIf
    \State $iter \gets iter + 1$
\Until{$ratio \leq \theta$ \textbf{or} $iter \geq max\_iter$}
\State \Return $T_r$ \Comment{Return final repainted template image}
\end{algorithmic}
\end{algorithm}

\section{Appendix C: Extended Quantitative Results}
Fig.\ref{fig:quality_eval} illustrates the performance across multiple iterations utilizing the Quality Evaluator. We limit the Anywhere framework to three iterations to balance efficiency and quality. Tab.\ref{tab:time} presents a comprehensive breakdown of time consumption for each component within the Anywhere framework, as well as for other inpainting models.

\setcounter{figure}{0}
\renewcommand{\thefigure}{A\arabic{figure}}
\begin{figure}[htbp]
\centering
\begin{minipage}[b]{0.45\textwidth}
    \centering
    \includegraphics[width=\textwidth]{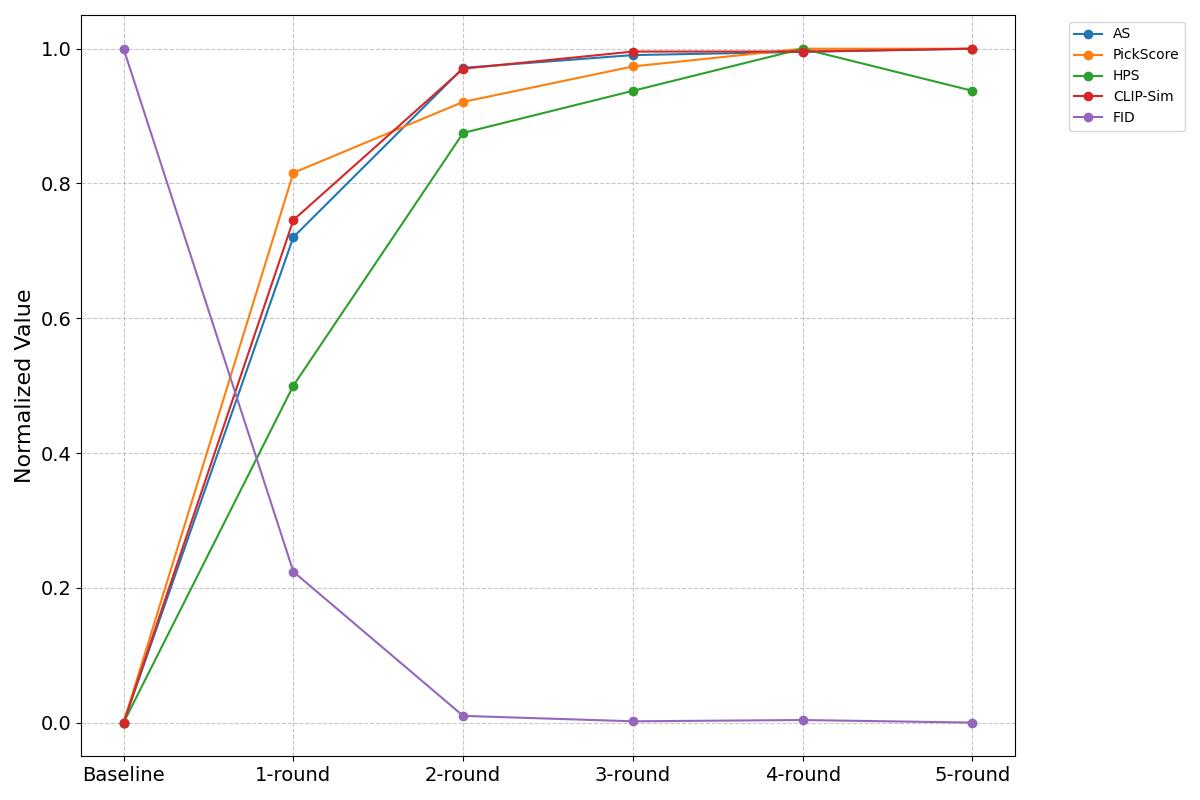}
    \caption{Impact of Quality Evaluator iterations on performance metrics. This graph illustrates the normalized change in various evaluation metrics across multiple rounds of Quality Evaluator feedback. Starting from the baseline (0 rounds), we show how these metrics evolve through 5 iterations, demonstrating the trade-off between quality improvement and computational cost. The results indicate that three iterations provide an optimal balance between performance gains and efficiency.}
    \label{fig:quality_eval}
\end{minipage}
\hfill
\begin{minipage}[b]{0.52\textwidth}
    \centering
    \begin{tabular}{lr}
    \hline
    \textbf{Module / Method} & \textbf{Time (s)} \\
    \hline
    Anyhwere & \textbf{27.5} \\
    \quad Prompt Generation Module & 19.0 \\
    \quad Image Generation & 7.5 \\
    \quad Template Generator & 5.0 \\
    \quad Template Repainter & 1.5 \\
    \quad Image Enhancer & 1.0 \\
    Quality Evaluator & 5.0 \\
    \hline
    Stable Diffusion 2.0 Inpainting & \textbf{5.0} \\
    \hline
    BrushNet   & \textbf{16.0} \\
    \hline
    HD-Painter & \textbf{21.0} \\
    \hline
    \end{tabular}
    \captionof{table}{Average time consumption per image generation for different modules and methods. For consistency in our evaluation, we standardized the inference steps to 50 for each result image generation. All tests were conducted on an NVIDIA A6000 GPU, with the reported times representing averages calculated from a substantial dataset of 3,000 foreground images, ensuring statistical reliability.}
    \label{tab:time}
\end{minipage}
\end{figure}

\section{Appendix D: Extended Qualitative Results} \label{appendix_qual}

In this Appendix, we present additional qualitative results to further elucidate the capabilities and efficacy of our Anywhere framework.
Fig.~\ref{fig:appendix_img2img} exhibits supplementary qualitative outcomes for the text-free scenarios (I2I).
We also provide comprehensive results for the text-guided scenarios (TI2I), underscoring our framework's proficiency in integrating user-specified textual inputs. For generic phrase text inputs, Fig.\ref{fig:appendix_sunset}, Fig.\ref{fig:appendix_beach}, Fig.\ref{fig:appendix_snow}, and Fig.\ref{fig:appendix_room} display the generated results for the user inputs ``sunset'', ``beach'', ``snow'', and ``room'', respectively. Fig.~\ref{fig:appendix_longtext} showcases the results generated from unique sentences tailored to each foreground image, highlighting the framework's capacity to process more complex, context-specific prompts.

To offer a more nuanced understanding of our framework's components, we include qualitative results from our ablation studies. Fig.\ref{fig:appendix_ablation_pg} presents the qualitative outcomes of the ablation experiment for the Prompt Generation Module, while Fig.\ref{fig:appendix_ablation_repainter} illustrates the results for the Template Repainter agent. The impact of the Image Enhancer agent is depicted in Fig.\ref{fig:appendix_ablation_refine}, and Fig.\ref{fig:appendix_ablation_oa} demonstrates the contribution of the Quality Evaluator agent.

\setcounter{figure}{1}
\renewcommand{\thefigure}{A\arabic{figure}}
\begin{figure}[h]
\centering
\includegraphics[width=0.9\linewidth]{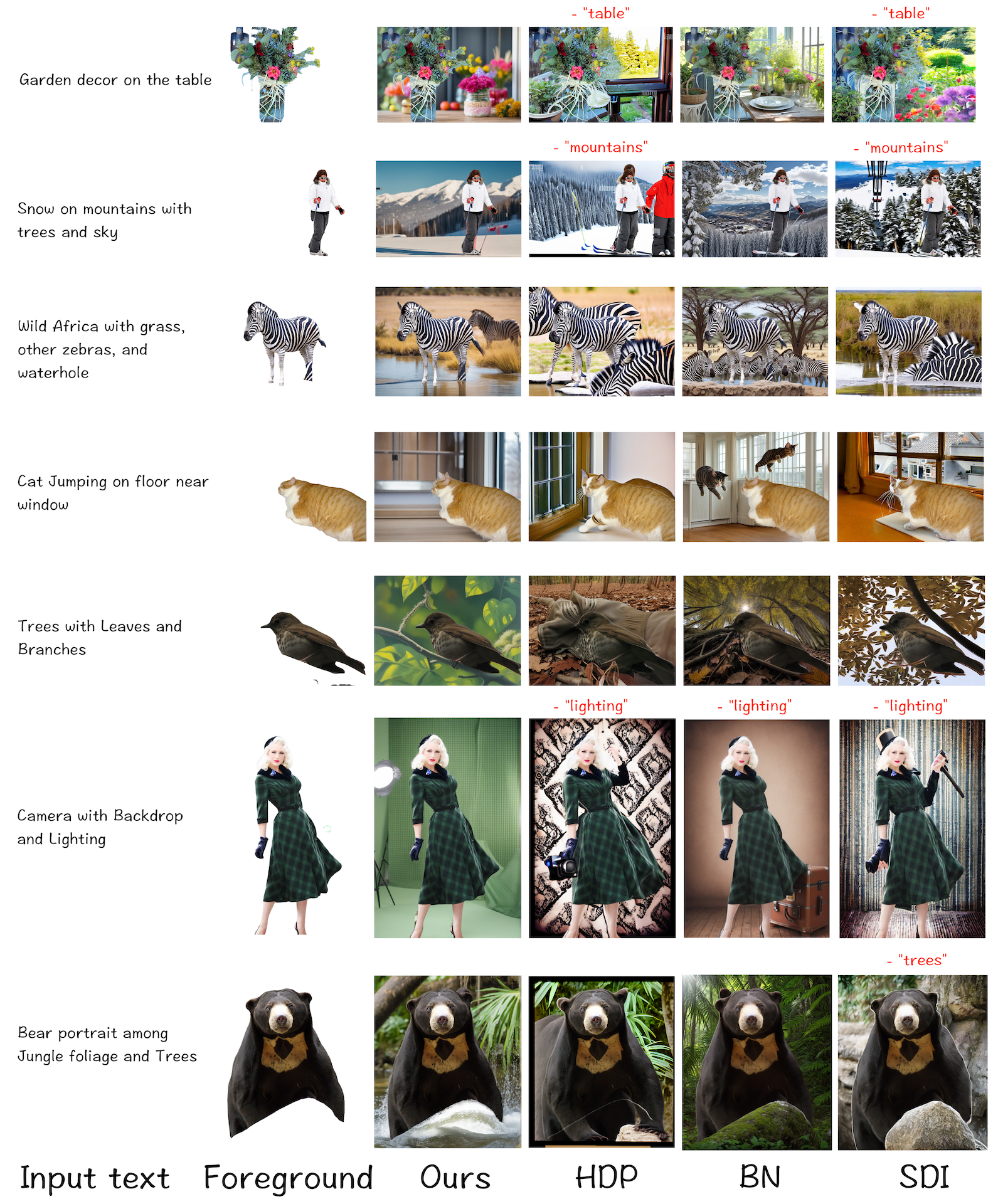}
\caption{More qualitative results on text-guided scenarios (TI2I).}\label{fig:appendix_longtext}
\end{figure}

\setcounter{figure}{2}
\renewcommand{\thefigure}{A\arabic{figure}}
\begin{figure}[h]
\centering
\includegraphics[width=0.9\linewidth]{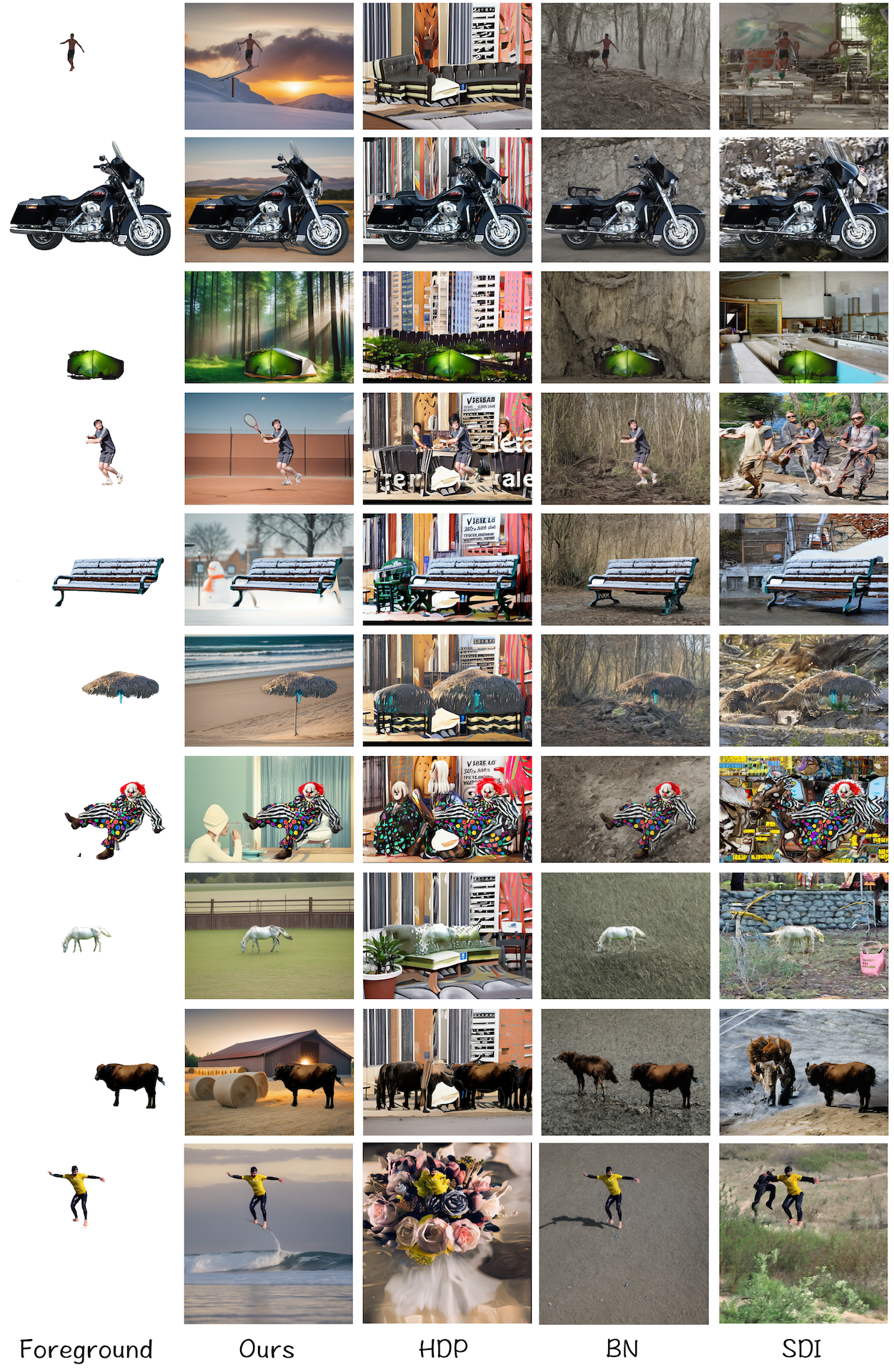}
\caption{More qualitative results on text-free scenarios (I2I).}\label{fig:appendix_img2img}
\end{figure}

\setcounter{figure}{3}
\renewcommand{\thefigure}{A\arabic{figure}}
\begin{figure}[h]
\centering
\includegraphics[width=0.7\linewidth]{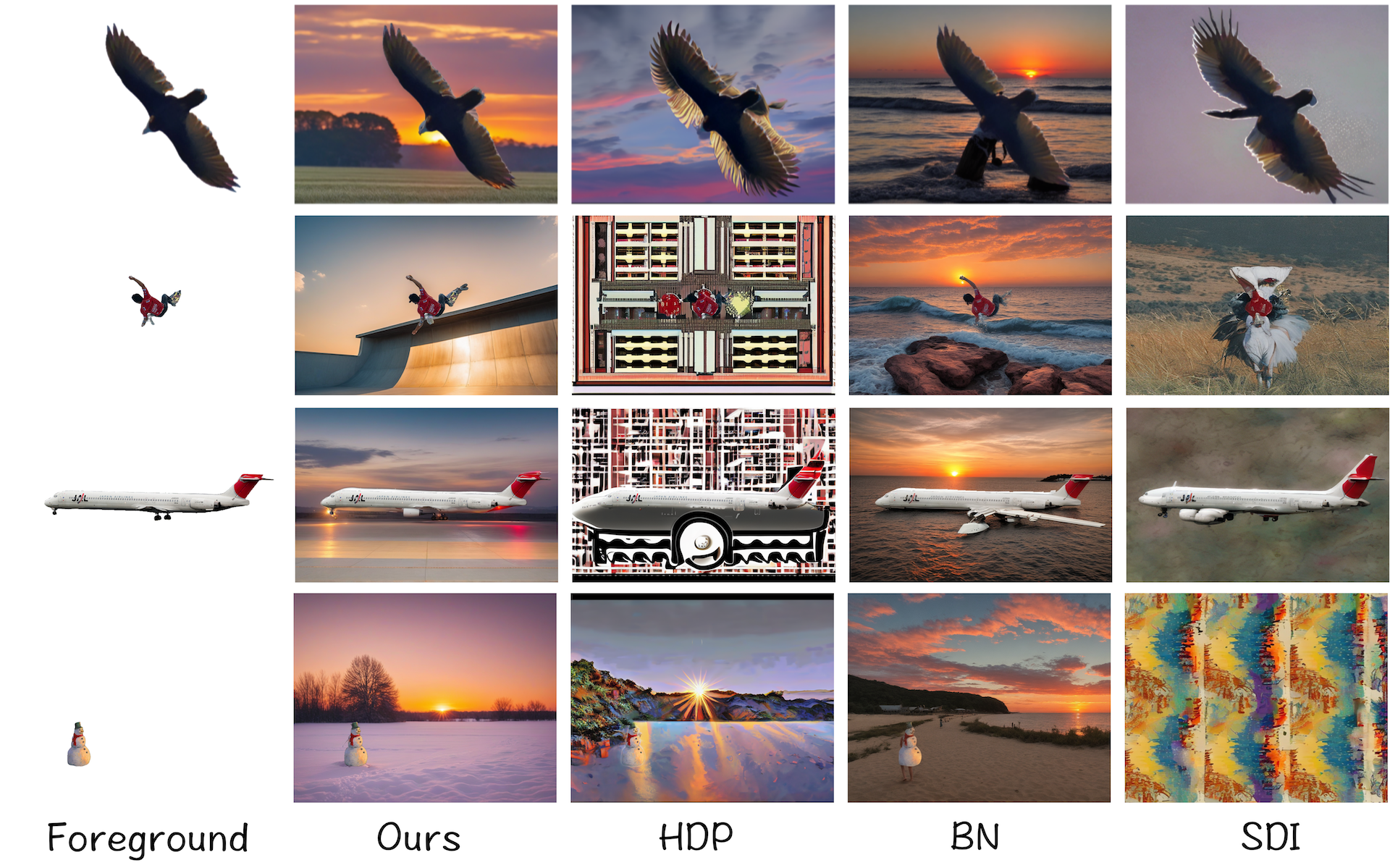}
\caption{More qualitative results on ``sunset'' user prompt.}\label{fig:appendix_sunset}
\end{figure}

\setcounter{figure}{4}
\renewcommand{\thefigure}{A\arabic{figure}}
\begin{figure}[h]
\centering
\includegraphics[width=0.7\linewidth]{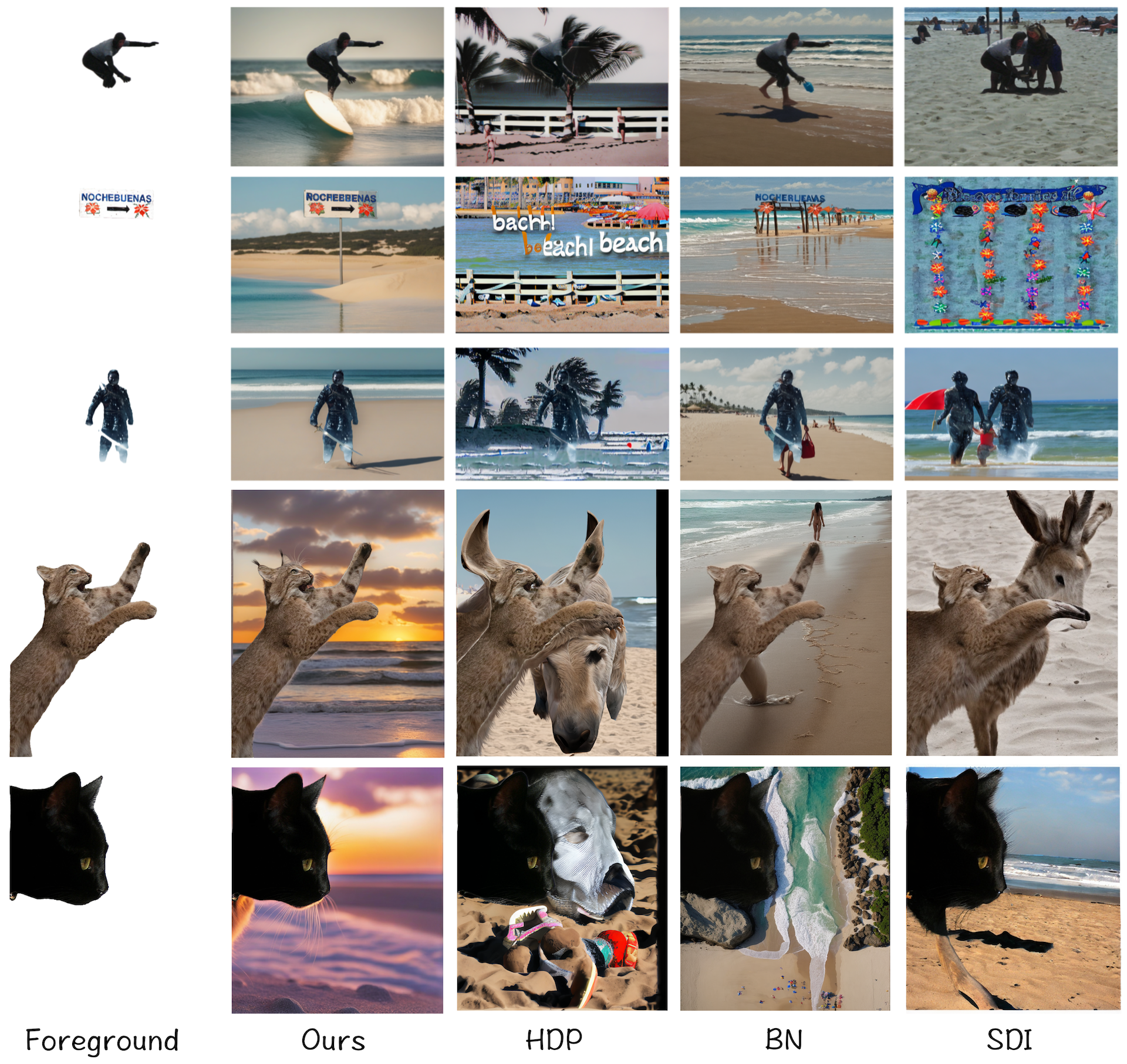}
\caption{Qualitative results on ``beach'' user prompt.}\label{fig:appendix_beach}
\end{figure}

\setcounter{figure}{5}
\renewcommand{\thefigure}{A\arabic{figure}}
\begin{figure}[h]
\centering
\includegraphics[width=0.65\linewidth]{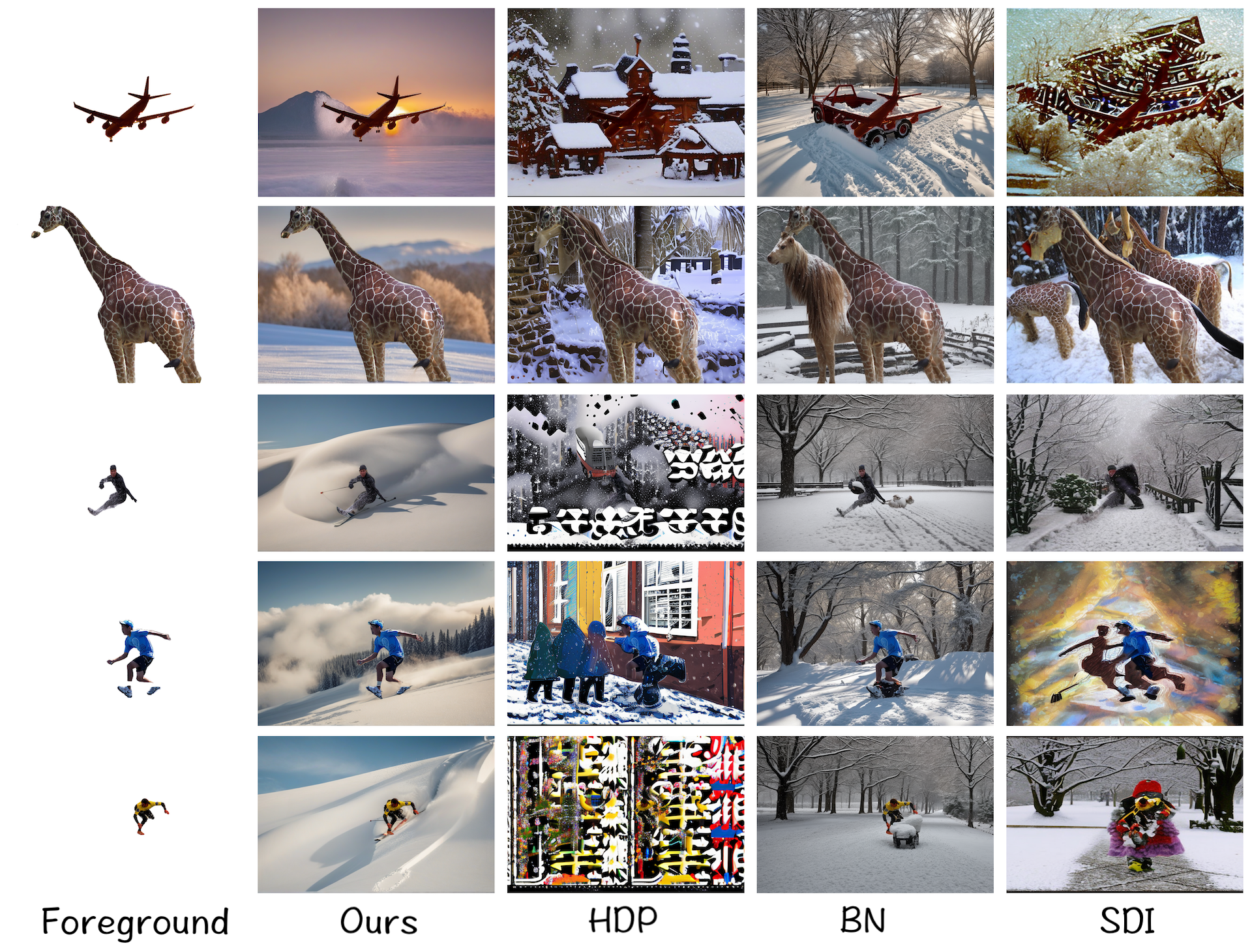}
\caption{Qualitative results on ``snow'' user prompt.}\label{fig:appendix_snow}
\end{figure}

\setcounter{figure}{6}
\renewcommand{\thefigure}{A\arabic{figure}}
\begin{figure}[h]
\centering
\includegraphics[width=0.65\linewidth]{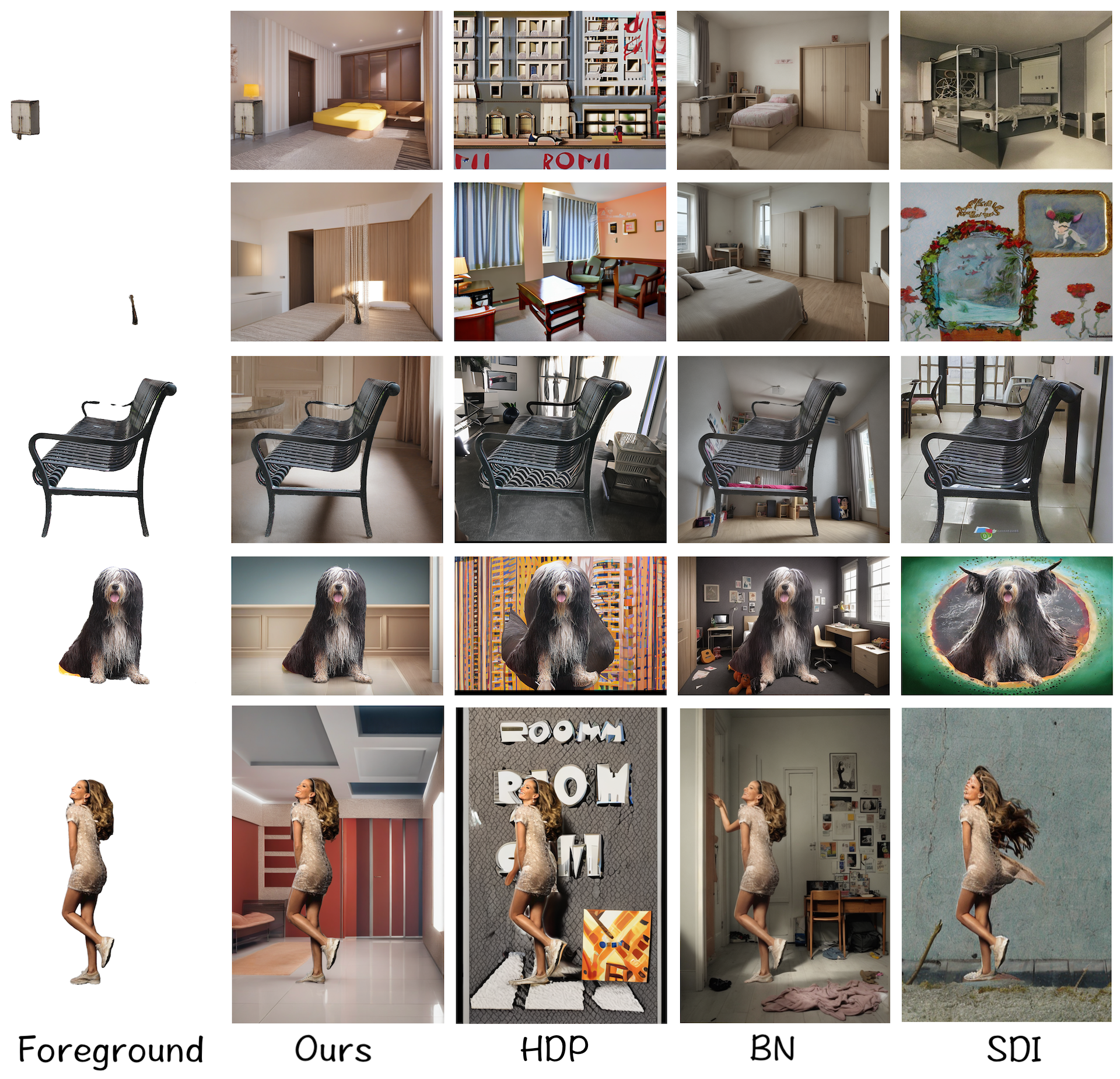}
\caption{Qualitative results on ``room'' user prompt.}\label{fig:appendix_room}
\end{figure}


\setcounter{figure}{7}
\renewcommand{\thefigure}{A\arabic{figure}}
\begin{figure}[htbp]
    \centering
    \begin{subfigure}[b]{0.48\textwidth}
        \centering
        \includegraphics[width=\textwidth]{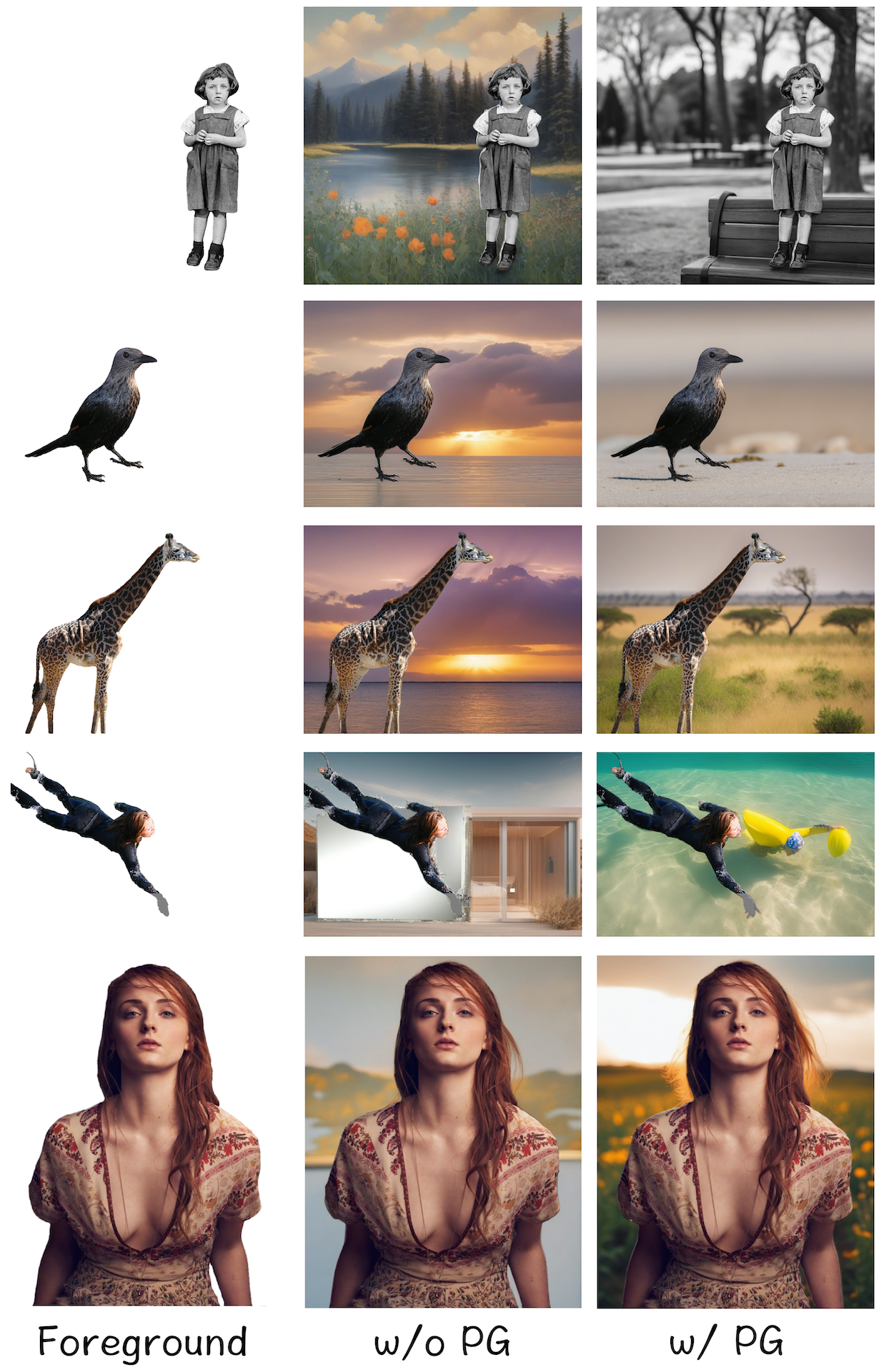}
        \caption{Qualitative results of ablation study on Prompt Generation module (PGM).}
        \label{fig:appendix_ablation_pg}
    \end{subfigure}
    \hfill
    \begin{subfigure}[b]{0.48\textwidth}
        \centering
        \includegraphics[width=\textwidth]{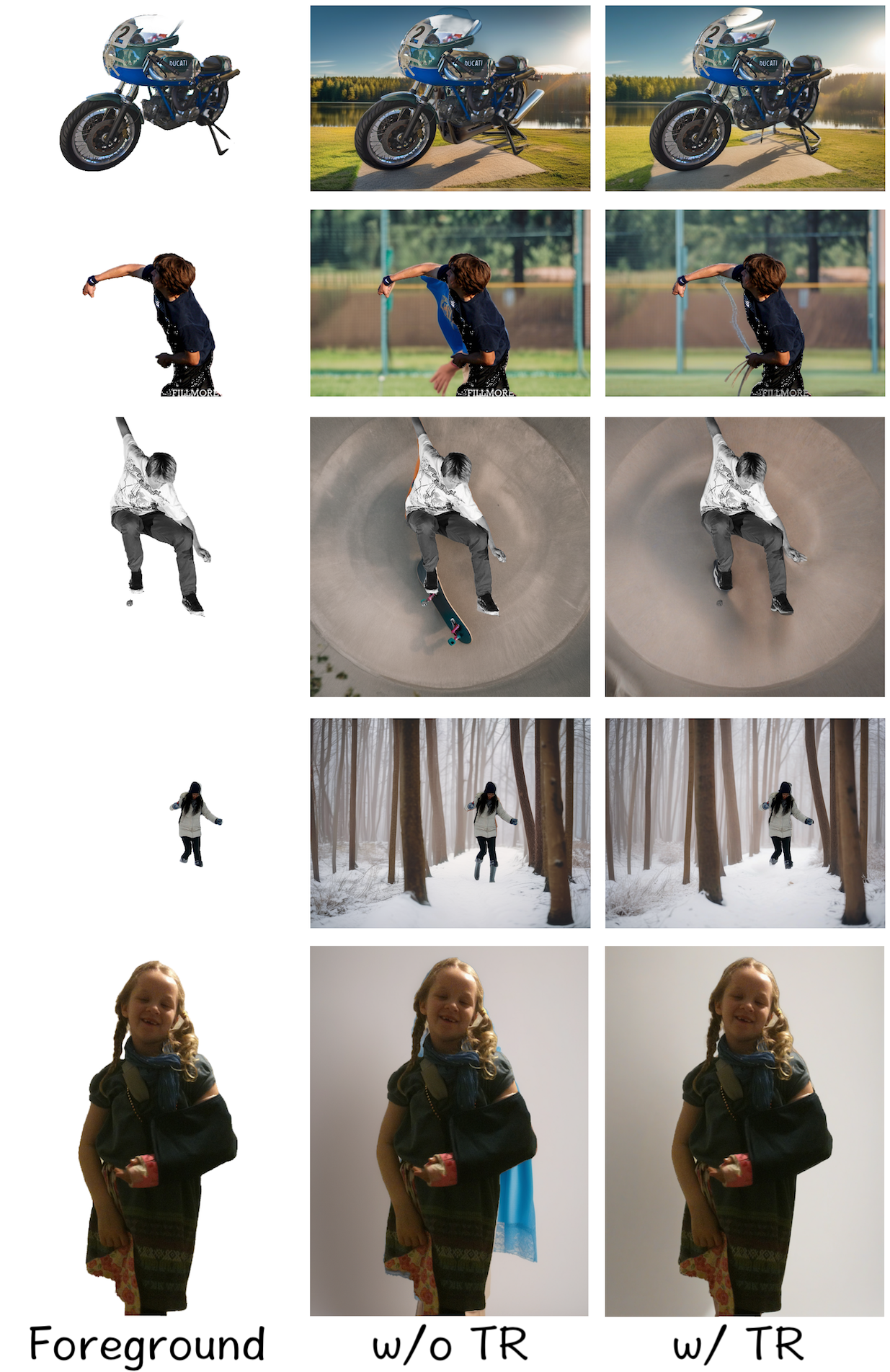}
        \caption{Qualitative results of ablation study on Template Repainter (TR).}
        \label{fig:appendix_ablation_repainter}
    \end{subfigure}
    \caption{Ablation studies on Prompt Generation module (PGM) and Template Repainter (TR).}
    \label{fig:combined_ablation_pg_repainter}
\end{figure}

\setcounter{figure}{8}
\renewcommand{\thefigure}{A\arabic{figure}}
\begin{figure}[htbp]
    \centering
    \begin{subfigure}[b]{0.45\textwidth}
        \centering
        \includegraphics[width=\textwidth]{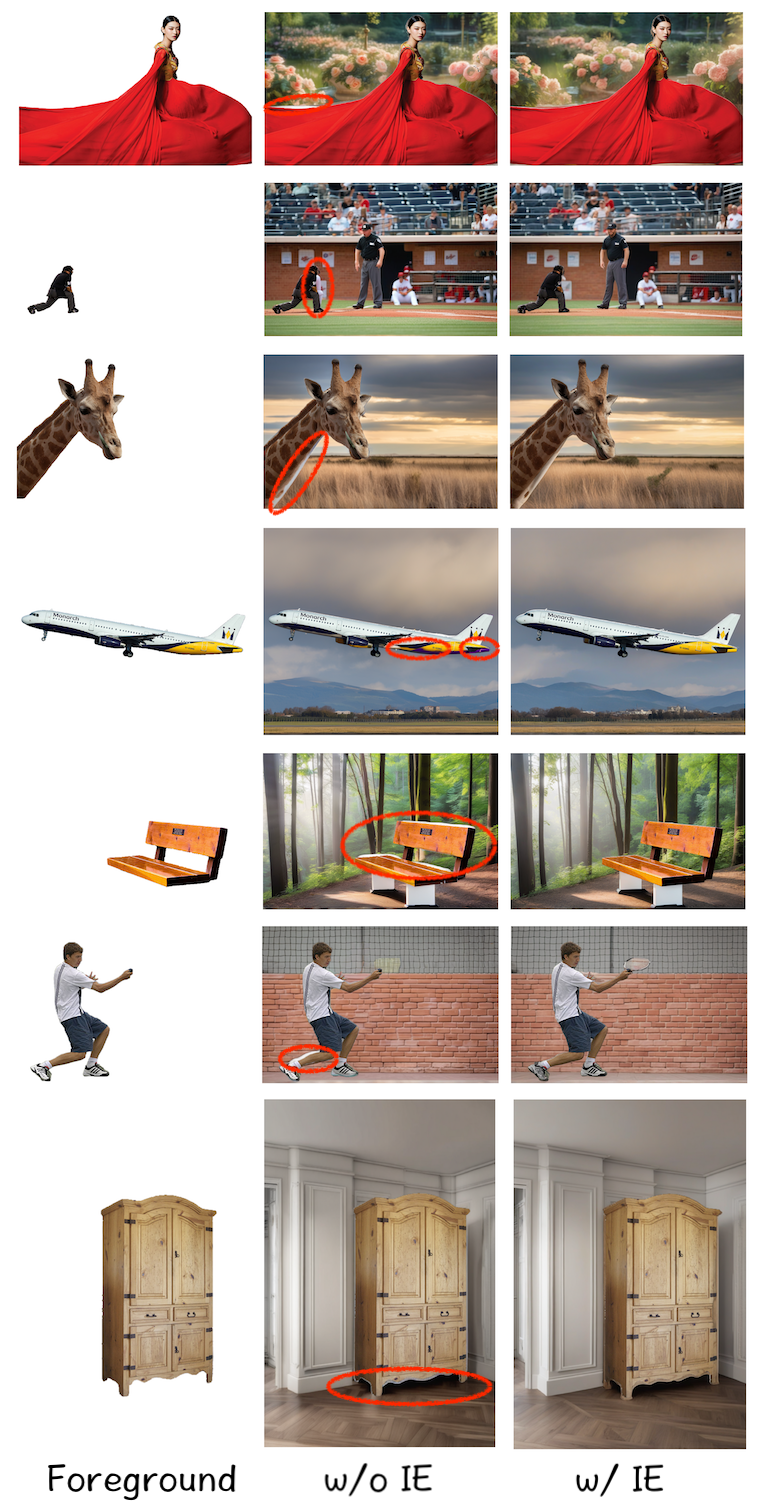}
        \caption{Qualitative results of ablation study on Image Enhancer (IE).}
        \label{fig:appendix_ablation_refine}
    \end{subfigure}
    \hfill
    \begin{subfigure}[b]{0.54\textwidth}
        \centering
        \includegraphics[width=\textwidth]{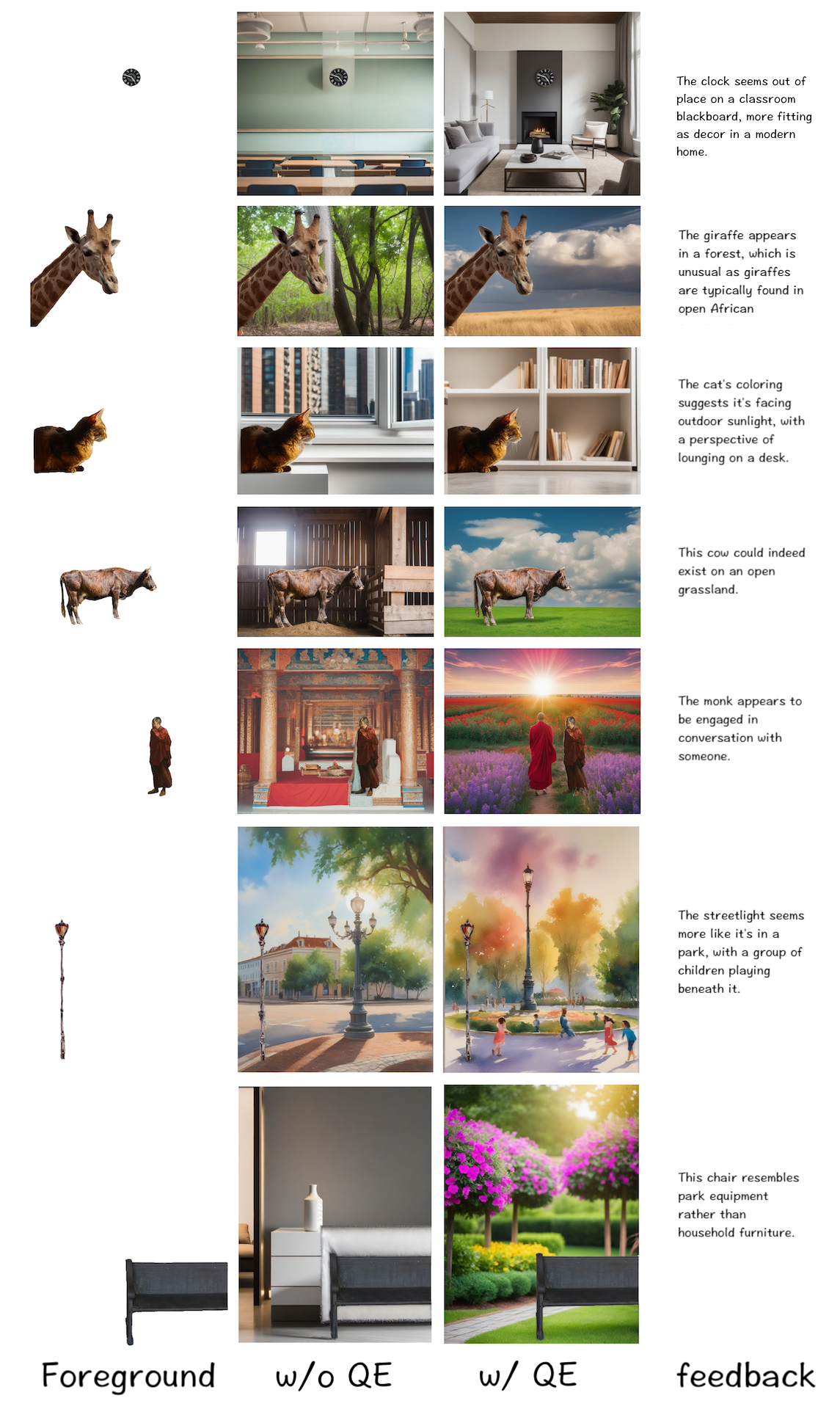}
        \caption{Qualitative results of ablation study on Quality Evaluator (QE).}
        \label{fig:appendix_ablation_oa}
    \end{subfigure}
    \caption{Ablation studies on Image Enhancer (IE) and Quality Evaluator (QE).}
    \label{fig:combined_ablation}
\end{figure}

\end{appendices}

\end{document}